\documentclass{article}

\usepackage{microtype}
\usepackage{graphicx}
\usepackage{subfigure}
\usepackage{booktabs} 
\usepackage[utf8]{inputenc}
\usepackage{url}
\usepackage{amsmath}

\usepackage{hyperref}

\usepackage{xcolor}

\usepackage[accepted]{sysml2019}

\sysmltitlerunning{Ludwig: a type-based declarative deep learning toolbox}

\begin{document}

    \twocolumn[
        \sysmltitle{Ludwig: a type-based declarative deep learning toolbox}
        
        
        
        \sysmlsetsymbol{equal}{*}
        
        \begin{sysmlauthorlist}
        \sysmlauthor{Piero Molino}{uberai}
        \sysmlauthor{Yaroslav Dudin}{uberai}
        \sysmlauthor{Sai Sumanth Miryala}{uberai}
        \end{sysmlauthorlist}
        
        \sysmlaffiliation{uberai}{Uber AI, San Francisco, USA}
        
        \sysmlcorrespondingauthor{Piero Molino}{piero@uber.com}
        
        \sysmlkeywords{Machine Learning, SysML}
        
        \vskip 0.3in
    
        \begin{abstract}
            In this work we present Ludwig, a flexible, extensible and easy to use toolbox which allows users to train deep learning models and use them for obtaining predictions without writing code.
            Ludwig implements a novel approach to deep learning model building based on two main abstractions: data types and declarative configuration files.
    		The data type abstraction allows for easier code and sub-model reuse, and the standardized interfaces imposed by this abstraction allow for encapsulation and make the code easy to extend.
    		Declarative model definition configuration files enable inexperienced users to obtain effective models and increase the productivity of expert users.
    		Alongside these two innovations, Ludwig introduces a general modularized deep learning architecture called Encoder-Combiner-Decoder that can be instantiated to perform a vast amount of machine learning tasks.
    		These innovations make it possible for engineers, scientists from other fields and, in general, a much broader audience to adopt deep learning models for their tasks, concretely helping in its democratization.
        \end{abstract}
    ]

    
    
    \printAffiliationsAndNotice{} 
	
	\section{Introduction and Motivation}
	
	Over the course of the last ten years, deep learning models have demonstrated to be highly effective in almost every machine learning task in different domains including (but not limited to) computer vision, natural language, speech, and recommendation.
	Their wide adoption in both research and industry have been greatly facilitated by increasingly sophisticated software libraries like Theano~\cite{2016arXiv160502688short}, TensorFLow~\cite{tensorflow2015-whitepaper}, Keras~\cite{chollet2015keras}, PyTorch~\cite{paszke2017automatic}, Caffe~\cite{jia2014caffe}, Chainer~\cite{tokui2015chainer}, CNTK~\cite{seide2016cntk} and MXNet~\cite{chen2015mxnet}.
	Their main value has been to provide tensor algebra primitives with efficient implementations which, together with the massively parallel computation available on GPUs, enabled researchers to scale training	to bigger datasets.
	Those packages, moreover, provided standardized implementations of automatic differentiation, which greatly simplified model implementation.
	Researchers, without having to spend time re-implementing these basic building blocks from scratch and now having fast and reliable implementations of the same, were able to focus on models and architectures, which led to the explosion of new *Net model architectures of the last five years.
	
	With artificial neural network architectures being applied to a wide variety of tasks, common practices regarding how to handle certain types of input information emerged.
	When faced with a computer vision problem, a practitioner pre-processes data using the same pipeline that resizes images, augments them with some transformation and maps them into 3D tensors.
	Something similar happens for text data, where text is tokenized either into a list of words or characters or word pieces, a vocabulary with associated numerical IDs is collected and sentences are transformed into vectors of integers.
	Specific architectures are adopted to encode different types of data into latent representations: convolutional neural networks are used to encode images and recurrent neural networks are adopted for sequential data and text (more recently self-attention architectures are replacing them).
	Most practitioners working on a multi-class classification task would project latent representations into vectors of the size of the number of classes to obtain logits and apply a \textit{softmax} operation to obtain probabilities for each class, while for regression tasks, they would map latent representations into a single dimension by a linear layer, and the single score is the predicted value.
	
	\begin{figure*}[ht]
		\centering
		\includegraphics[width=0.8\linewidth]{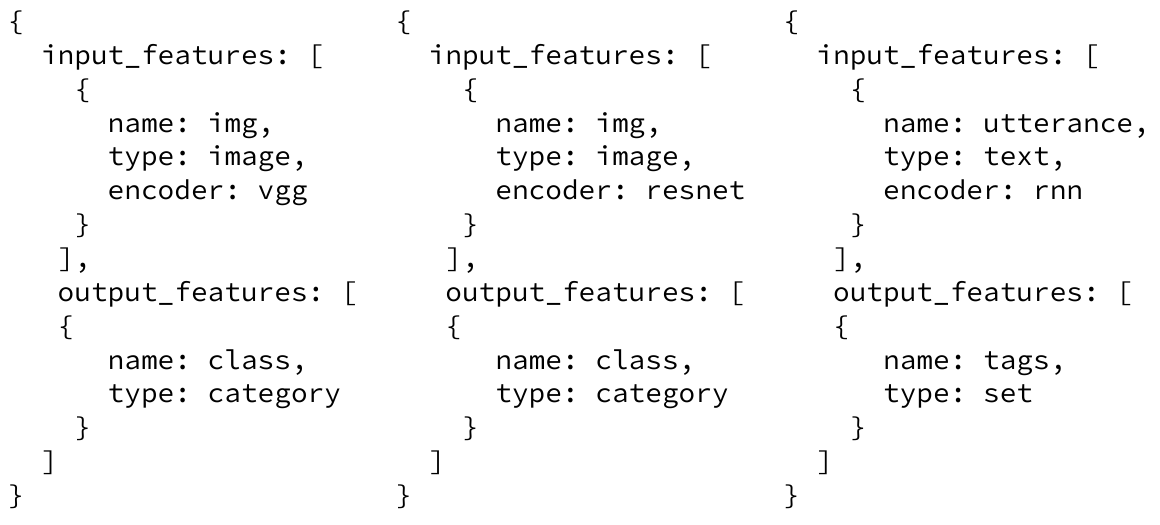}
		\caption{Examples of declarative model definitions. The first two show two models for image classification using two different encoders, while the third shows a multi-label text classification system. Note that a part from the name of input and output features, which are just identifiers, all that needs to be changed to encode with a different image encoder is jest the name of the encoder, while for changing tasks all that needs to be changed is the types of the inputs and outputs.}
		\label{fig:model_definition_examples}
	\end{figure*}
	
	Observing these emerging patterns led us to define abstract functions that identify classes of equivalence of model architectures.
	For instance, most of the different architectures for encoding images can be seen as different implementations of the abstract encoding function $T'_{h' \times w' \times c'} = e_\theta(T_{h \times w \times c})$ where $T_{dims}$ denotes a tensor with dimensions $dims$ and $e_\theta$ is an encoding function parametrized by parameters $\theta$ that maps from tensor to tensor.
	In tasks like image classification, $T'$ is pooled and flattened (i.e., a reduce function is applied spatially and the output tensor is reshaped as a vector) before being provided to, again, an abstract function that computes $T'_c=d_\theta(T_h)$ where $T_h$ is a one-dimensional tensor of hidden size $h$, $T'_c$ is a one-dimensional tensor of size $c$ equal to the number of classes, and $d_\theta$ is a decoding function parametrized by parameters $\theta$ that maps a hidden representation into logits and is usually implemented as a stack of fully connected layers.
	Similar abstract encoding and decoding functions that generalize many different architectures can be defined for different types of input data and different types of expected output predictions (which in turn define different types of tasks). 
	
	We introduce Ludwig, a deep learning toolbox based on the above-mentioned level of abstraction, with the aim to encapsulate best practices and take advantage of inheritance and code modularity.
	Ludwig makes it much easier for practitioners to compose their deep learning models by just declaring their data and task and to make code reusable, extensible and favor best practices.
	These classes of equivalence are named after the data type of the inputs encoded by the encoding functions (image, text, series, category, etc.) and the data type of the outputs  predicted by the decoding functions.
	This type-based abstraction allows for a higher level interface than what is available in current deep learning frameworks, which abstract at the level of single tensor operation or at the layer level.
	This is achieved by defining abstract interfaces for each data type, which allows for extensibility as any new implementation of the interface is a drop-in replacement for all the other implementations already available.
	
	Concretely, this allows for defining, for instance, a model that includes an image encoder and a category decoder and being able to swap in and out VGG~\cite{DBLP:journals/corr/SimonyanZ14a},  ResNet~\cite{he2016deep} or DenseNet~\cite{Huang_2017_CVPR} as different interchangeable representations of an image encoder.
	The natural consequence of this level of abstraction is associating a name to each encoder for a specific type and enabling the user to declare what model to employ rather than requiring them to implement them imperatively, and at the same time, letting the user add new and custom encoders.
	The same also applies to data types other than images and decoders.
	
	With such type-based interfaces in place and implementations of such interfaces readily available, it becomes possible to construct a deep learning model simply by specifying the type of the features in the data and selecting the implementation to employ for each data type involved.
	Consequently, Ludwig has been designed around the idea of a declarative specification of the model to allow a much wider audience (including people who do not code) to be able to adopt deep learning models, effectively democratizing them.
	Three such model definition are shown in Figure~\ref{fig:model_definition_examples}.
	
	The main contribution of this work is that, thanks to this higher level of abstraction and its declarative nature, Ludwig allows for inexperienced users to easily build deep learning models
	, while allowing experts to decide the specific modules to employ with their hyper-parameters and to add additional custom modules.
	Ludwig's other main contribution is the general modular architecture defined through the type-based abstraction that allows for code reuse, flexibility, and the performance of a wide array of machine learning tasks under a cohesive framework.
	
	The remainder of this work describes Ludwig's architecture in detail, explains its implementation, compares Ludwig with other deep learning frameworks and discusses its advantages and disadvantages.

	\section{Architecture}
	The notation used in this section is defined as follows.
	Let $d \sim D$ be a data point sampled from a dataset $D$.
	Each data point is a tuple of typed values called features.
	They are divided in two sets: $d_I$ is the set of input features and $d_O$ id the set of output features.
	$d_i$ will refer to a specific input feature, while $d_o$ will refer to a specific output features.
	Model predictions given input features $d_I$ are denoted as $d_P$, so that there will be a specific prediction $d_p$ for each output feature $d_o \in d_O$.
	The types of the features can be either atomic (scalar) types like binary, numerical or category, or complex ones like discrete sequences or sets.
    Each data type is associated with abstract function types, as is explained in the following section, to perform type-specific operations on features and tensors.
    Tensors are a generalization of scalars, vectors, and matrices with $n$ ranks of different dimensions.
    Tensors are referred to as $T_{dims}$ where $dims$ indicates the dimensions for each rank, like for instance $T_{l \times m \times n}$ for a rank 3 tensor of dimensions $l$, $m$ and $n$ respectively for each rank.

	\subsection{Type-based Abstraction}
	
	\begin{figure}
		\centering
		\includegraphics[width=\columnwidth]{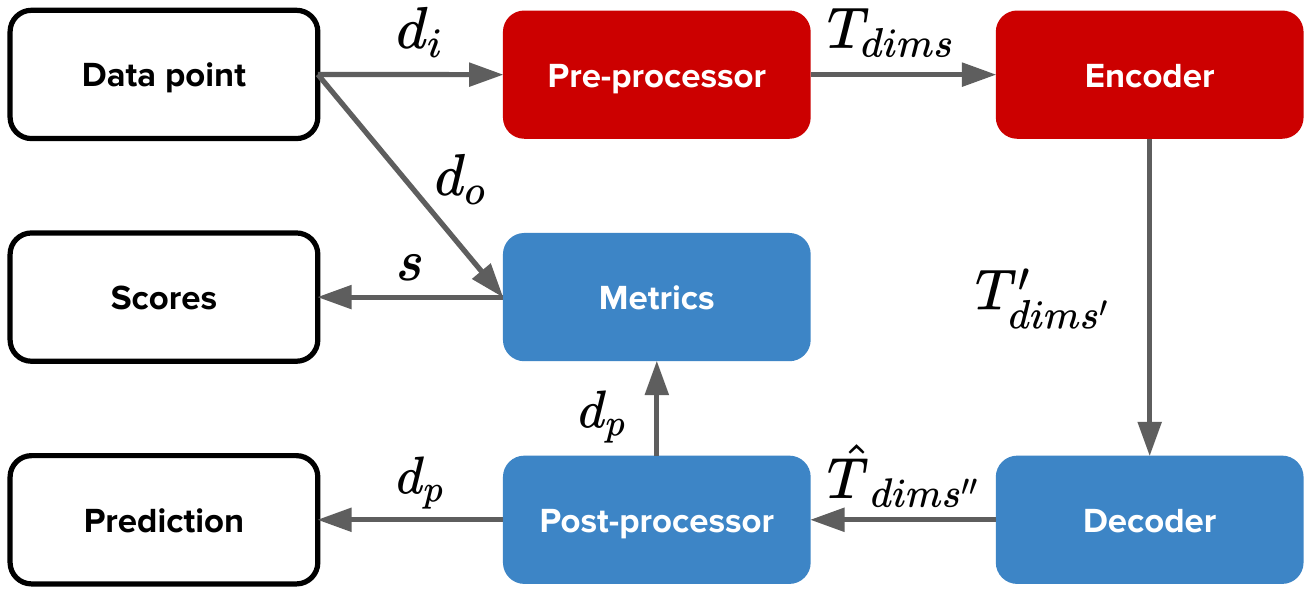}
		\caption{Data type functions flow.}
		\label{fig:type_flow}
	\end{figure}
	
	Type-based abstraction is one of the main concepts that define Ludwig's architecture.
	Currently, Ludwig supports the following types: binary, numerical (floating point values), category (unique strings), set of categorical elements, bag of categorical elements, sequence  of categorical elements, time series (sequence of numerical elements), text, image, audio (which doubles as speech when using different pre-processing parameters), date, H3~\cite{brodsky2018} (a geo-spatial indexing system), and vector (one dimensional tensor of numerical values).
	The type-based abstraction makes it easy to add more types.
	
	The motivation behind this abstraction stems from the observation of recurring patterns in deep learning projects: pre-processing code is more or less the same given certain types of inputs and specific tasks, as is the code implementing models and training loops.
	Small differences make models hard to compare and their code difficult to reuse.
	By modularizing it on a data type base, our aim is to improve both code reuse, adoption of best practices and extensibility.

	\begin{figure*}[!ht]
		\centering
		\includegraphics[width=0.7\linewidth]{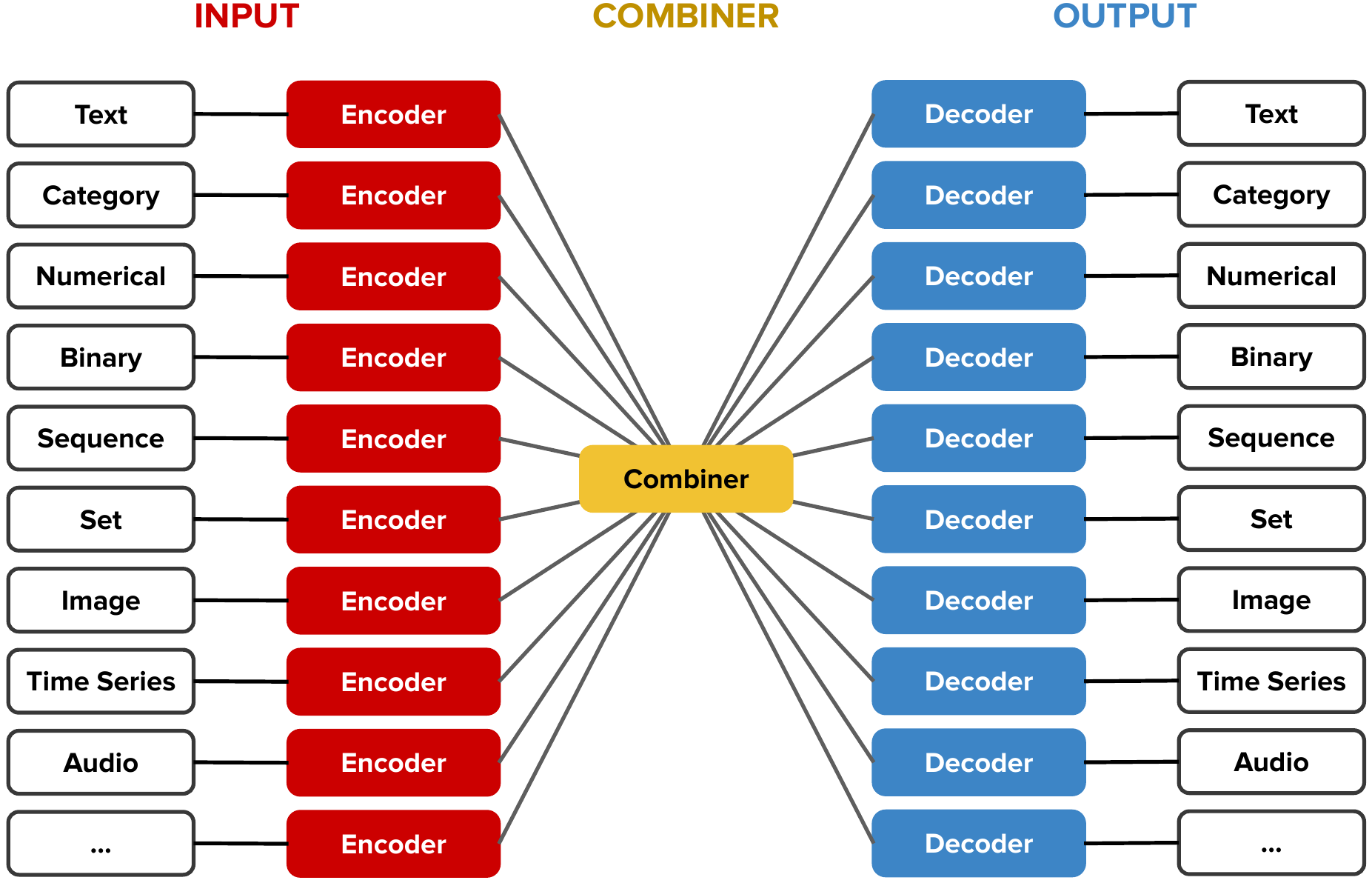}
		\caption{Encoder-Combiner-Decoder Architecture}
		\label{fig:ECD}
	\end{figure*}
	
	Each data type has five abstract function types associated with it and there could be multiple implementations of each of them:
	
	\begin{itemize}
		
		\item \textbf{Pre-processor}: a pre-processing function $T_{dims}=\mathrm{pre}(d_i)$ maps a raw data point input feature $d_i$ into a tensor $T$ with dimensions $dims$.
		Different data types may have different pre-processing functions and different dimensions of $T$.
		A specific type may, moreover, have different implementations of $\mathrm{pre}$. A concrete example is \textit{text}: $d_i$ in this case is a string of text, there could be different tokenizers that implement $\mathrm{pre}$ by splitting on space or using byte-pair encoding and mapping tokens to integers, and $dims$ is $s$, the length of the sequence of tokens.
		
		\item \textbf{Encoder}: an encoding function $T'_{dims'} = e_\theta(T_{dims})$ maps an input tensor $T$ into an output tensor $T'$ using parameters $\theta$.
		The dimensions $dims$ and $dims'$ may be different from each other and depend on the specific data type. The input tensor is the output of a $\mathrm{pre}$ function. Concretely, encoding functions for \textit{text}, for instance, take as input $T_s$ and produce $T_h$ where $h$ is an hidden dimension if the output is required to be pooled, or $T_{s \times h}$ if the output is not pooled. Examples of possible implementations of $e_\theta$ are CNNs, bidirectional LSTMs or Transformers.
		
		\item \textbf{Decoder}: a decoding function $\hat{T}_{dims''} = d_\theta(T'_{dims'})$ maps an input tensor $T'$ into an output tensor $\hat{T}$ using parameter $\theta$. The dimensions $dims''$ and $dims'$ may be different from each other and depend on the specific data type. $T'$ is the output of an encoding function or of a combiner (explained in the next section). Concretely, a decoder function for the \textit{category} type would map $T_h$ input tensor into a $T_c$ tensor where $c$ is the number of classes.
		
		\item \textbf{Post-processor}: a post-processing function $d_p=\mathrm{post}(\hat{T}_{dims''})$ maps a tensor $\hat{T}$ with dimensions $dims''$ into a raw data point prediction $d_p$. $\hat{T}$ is the output of a decoding function. Different data types may have different post-processing functions and different dimensions of $T$. A specific type may, moreover, have different implementations of $\mathrm{post}$. A concrete example is \textit{text}: $d_p$ in this case is a string of text, and there could be different functions that implement $\mathrm{post}$ by first mapping integer predictions into tokens and then concatenating on space or using byte-pair concatenation to obtain a single string of text.

		\item \textbf{Metrics}:
		a metric function $s=m(d_o, d_p)$ produces a score $s$ given a ground truth output feature $d_o$ and predicted output $d_p$ of the same dimension. $d_p$ is the output of a post-processing function. In this context, for simplicity, loss functions are considered to belong to the metric class of function. Many different metrics may be associated with the same data type. Concrete examples of metrics for the \textit{category} data type can be accuracy, precision, recall, F1, and cross entropy loss, while for the \textit{numerical} data type they could be mean squared error, mean absolute error, and R2.
		
	\end{itemize}
	
	A depiction of how the functions associated with a data type are connected to each other is provided in Figure~\ref{fig:type_flow}.

	\subsection{Encoders-Combiner-Decoders} \label{sub:ECD}
	
	\begin{figure*}[ht]
		\centering
		\includegraphics[width=\linewidth]{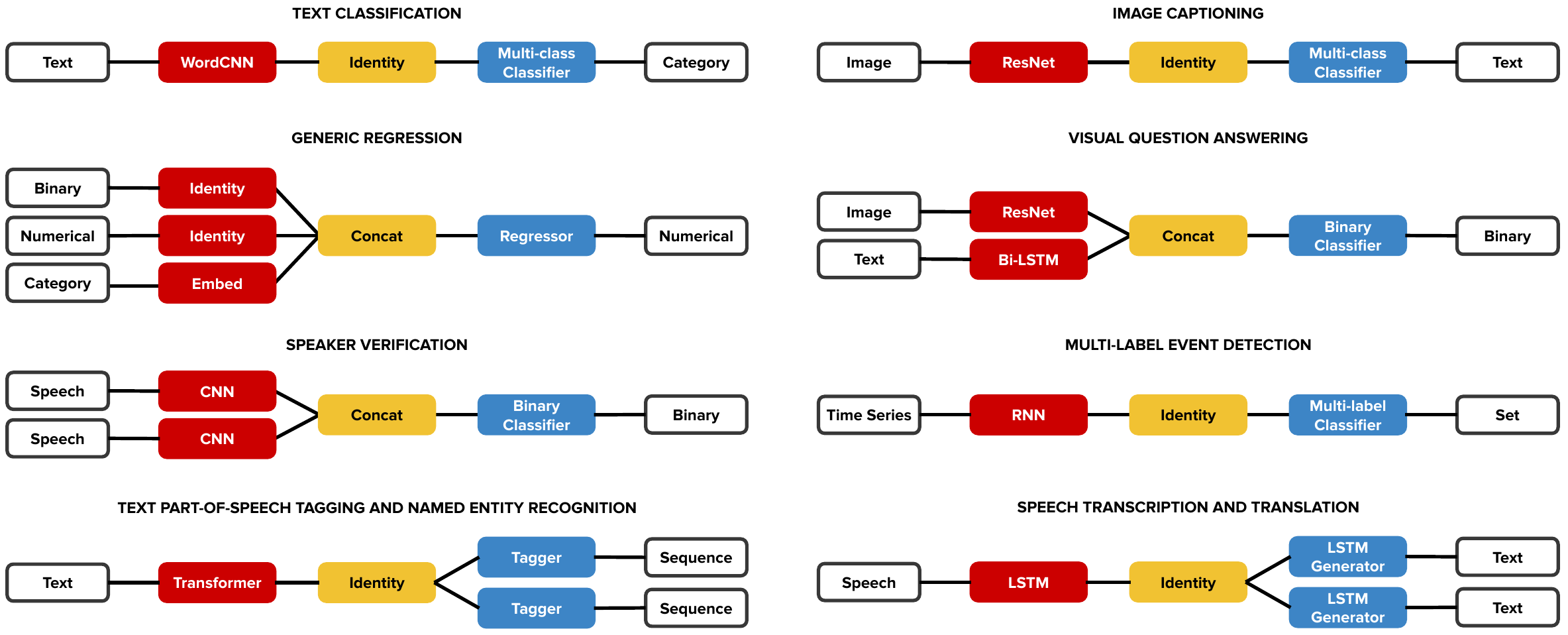}
		\caption{Different instantiations of the \textit{ECD} architecture for different machine learning tasks}
		\label{fig:ECD_instatiations}
	\end{figure*}
	
	In Ludwig, every model is defined in terms of encoders that encode different features of an input data point, a combiner which combines information coming from the different encoders, and decoders that decode the information from the combiner into one or more output features.
	This generic architecture is referred to as Encoders-Combiner-Decoders (ECD).
	A depiction is provided in Figure~\ref{fig:ECD}.
	
	This architecture is introduced because it maps naturally most of the architectures of deep learning models and allows for modular composition.
	This characteristic, enabled by the data type abstraction, allows for defining models by just declaring the data types of the input and output features involved in the task and assembling standard sub-modules accordingly rather than writing a full model from scratch.
	
	A specific instantiation of an \textit{ECD} architecture can have multiple input features of different or same type, and the same is true for output features.
	For each feature in the input part, pre-processing and encoding functions are computed depending on the type of the feature, while for each feature in the output part, decoding, metrics and post-processing functions are computed, again depending on the type of each output feature. 
	
	When multiple input features are provided a combiner function $\{T''\}=c_\theta({T'})$ that maps a set of input tensors $\{T'\}$ into a set of output tensors $\{T''\}$ is computed.
	$c$ has an abstract interface and many different functions can implement it.
	One concrete example is what in Ludwig is called \textit{concat} combiner: it flattens all the tensors in the input set, concatenates them and passes them to a stack of fully connected layers, the output of which is provided as output, a set of only one tensor.
	Note that a possible implementation of a combiner function can be the identity function. 
	
	This definition of a decoder function allows for implementations where subsets of inputs are provided to different sub-modules which return subsets of the output tensors, or even for a recursive definition where the combiner function is a \textit{ECD} model itself, albeit without pre-processors and post-processors, since inputs and outputs are already tensors and do not need to be pre-processed and post-processed.
	Although the combiner definition in the \textit{ECD} architecture is theoretically flexible, the current implementations of combiner functions in Ludwig are monolithic (without sub-modules), non-recursive, and return a single tensor as output instead of a set of tensors.
	However, more elaborate combiners can be added easily.
	
	The \textit{ECD} architecture allows for many instantiations by combining different input features of different data types with different output features of different data types, as depicted in Figure~\ref{fig:ECD_instatiations}.
	An \textit{ECD} with an input text feature and an output categorical feature can be trained to perform text classification or sentiment analysis, and an \textit{ECD} with an input image feature and a text output feature can be trained to perform image captioning, while an \textit{ECD} with categorical, binary and numerical input features and a numerical output feature can be trained to perform regression tasks like predicting house pricing, and an \textit{ECD} with numerical binary and categorical input features and a binary output feature can be trained to perform tasks like fraud detection.
	It is evident how this architecture is really flexible and is limited only by the availability of data types and the implementations of their functions. 
	
	\begin{figure*}
		\centering
		\includegraphics[width=\linewidth]{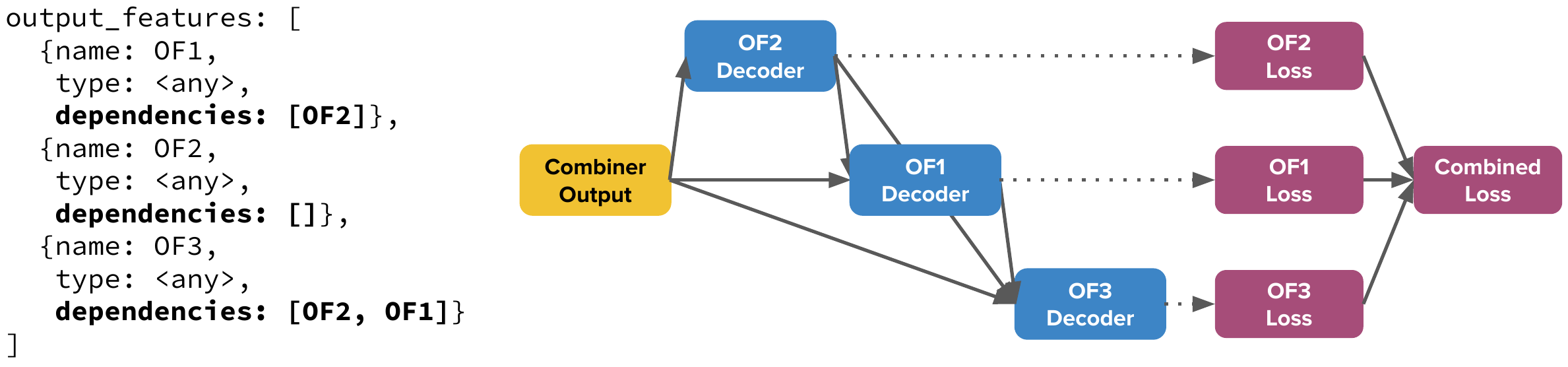}
		\caption{Different instantiations of the \textit{ECD} architecture for different machine learning tasks}
		\label{fig:output_feature_dependency}
	\end{figure*}
	An additional advantage of this architecture is its ability to perform multi-task learning~\cite{DBLP:conf/icml/Caruana93}.
	If more than one output feature is specified, an \textit{ECD} architecture can be trained to minimize the weighted sum of the losses of each output feature in an end-to-end fashion.
	This approach has shown to be highly effective in both vision and natural language tasks, achieving state of the art performance~\cite{DBLP:conf/cidr/RatnerHR19}.
	Moreover, multiple outputs can be correlated or have logical or statistical dependency with each other.
	For example, if the task is to predict both parts of speech and named entity tags from a sentence, the named entity tagger will most likely achieve higher performance if it is provided with the predicted parts of speech (assuming the predictions are better than chance, and there is correlation between part of speech and named entity tag).
	In Ludwig, dependencies between outputs can be specified in the model definition, a directed acyclic graph among them is constructed at model building time, and either the last hidden representation or the predictions of the origin output feature are provided as inputs to the decoder of the destination output feature.
	This process is depicted in Figure~\ref{fig:output_feature_dependency}.
	When non-differentiable operations are performed to obtain the predictions, for instance, like \textit{argmax} in the case of category features performing multi-class classification, the logits or the probabilities are provided instead, keeping the multi-task training process end-to-end differentiable.
	
	This generic formulation of multi-task learning as a directed acyclic graph of task dependencies is related to the hierarchical multi-task learning in Snorkel MeTaL proposed by \citet{DBLP:conf/sigmod/RatnerHDGR18} and its adoption for improving training from weak supervision by exploiting task agreements and disagreements of different labeling functions~\cite{DBLP:conf/aaai/RatnerHDSPR19}.
	The main difference is that Ludwig can handle automatically heterogeneous tasks, i.e. tasks to predict different data types with support for different decoders, while in Snorkel MeTaL each task head is a linear layer.
	On the other hand Snorkel MeTaL's focus on weak supervision is currently absent in Ludwig.
	An interesting avenue of further research to close the gap between the two approaches could be to infer dependencies and loss weights automatically given fully supervised multi-task data and combine weak supervision with heterogeneous tasks.
	
	\section{Implementation}
	
	\subsection{Declarative Model Definition}
	
	Ludwig adopts a declarative model definition schema that allows users to define an instantiation of the \textit{ECD} architecture to train on their data.
	
	The higher level of abstraction provided by the type-based \textit{ECD} architecture allows for a separation between what a model is expected to learn to do and how it actually does it.
	This convinced us to provide a declarative way of defining the models in Ludwig, as the amount of potential users who can define a model by declaring the inputs they are providing and the predictions they are expecting, without specifying the implementation of how the predictions are obtained, is substantially bigger than the amount of developers who can code a full deep learning model on their own.
	An additional motivation for the adoption of a declarative model definitions stems from the separation of interests between the authors of the implementations of the models and the final users, analogous to the separation of interests of the authors of query planning and indexing strategies of a database and those users who query the database, which allows the former to provide improved strategies without impacting the way the latter interacts with the system.
	
	\begin{figure*}
		\centering
		\includegraphics[width=\linewidth]{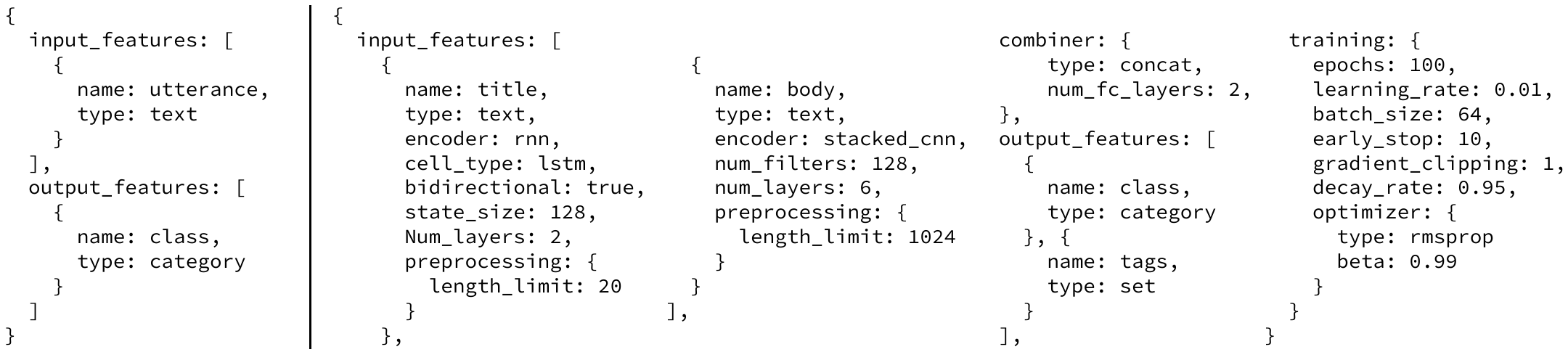}
		\caption{On the left side, a minimal model definition for text classification. On the right side, a more complex model definition including input and output features and more model and training hyper-parameters.}
		\label{fig:simple_complex_models}
	\end{figure*}
	
	The model definition is divided in five sections:
	
	\begin{itemize}
		
		\item \textbf{Input Features}: in this section of the model definition, a list of input features is specified. The minimum amount of information that needs to be provided for each feature is the name of the feature that corresponds to the name of a column in the tabular data provided by the user, and the type of such feature. Some features have multiple encoders, but if one is not specified, the default one is used. Each encoder can have its own hyper-parameters, and if they are not specified, the default hyper-parameters of the specified encoder are used. 
		
		\item \textbf{Combiner}: in this section of the model definition, the type of combiner can be specified, if none is specified, the default \textit{concat} is used. Each combiner can have its own hyper-parameters, but if they are not specified, the default ones of the specified combiner are used.
		
		\item \textbf{Output Features}: in this section of the model definition, a list of output features is specified. The minimum amount of information that needs to be provided for each feature is the name of the feature that corresponds to the name of a column in the tabular data provided by the user, and the type of such feature. The data in the column is the ground truth the model is trained to predict. Some features have multiple decoders that calculate the predictions, but if one is not specified, the default one is used. Each decoder can have its own hyper-parameters and if they are not specified, the default hyper-parameters of the specified encoder are used. Moreover, each decoder can have different losses with different parameters to compare the ground truth values and the values predicted by the decoder and, also in this case, if they are not specified, defaults are used.
		
		\item \textbf{Pre-processing}: pre-processing and post-processing functions of each data type can have parameters that change their behavior. They can be specified in this section of the model definition and are applied to all input and output features of a specified type, and if they are not provided, defaults are used. Note that for some use cases it would be useful to have different processing parameters for different features of the same type. Consider a news classifier where the title and the body of a piece of news are provided as two input text features. In this case, the user may be inclined to set a smaller value for the maximum length of words and the maximum size of the vocabulary for the title input feature. Ludwig allows users to specify processing parameters on a per-feature basis by providing them inside each input and output feature definition. If both type-level parameters and single-feature-level parameters are provided, the single-feature-level ones override the type-level ones.
		
		\item \textbf{Training}: the training process itself has parameters that can be changed, like the number of epochs, the batch size, the learning rate and its scheduling, and so on. Those parameters can be provided by the user, but if they are not provided, defaults are used.
		
	\end{itemize}
	
	The wide adoption of defaults allows for really concise model definitions, like the one shown on the left side of Figure~\ref{fig:simple_complex_models}, as well as a high degree of control on both the architecture of the model and training parameters, as shown on the right side of Figure~\ref{fig:simple_complex_models}.
	
	Ludwig adopts the convention to adopt YAML to parse model definitions because of its human readability, but as long its nested structure is representable, other similar formats could be adopted.

	For the ever-growing list of available encoders, combiners, and decoders, their hyper-parameters, the pre-processing and training parameter available, please consult Ludwig's user guide\footnote{\url{http://ludwig.ai/user_guide/}}. For additional examples refer to the example\footnote{\url{http://ludwig.ai/examples}} section.
	
	In order to allow for flexibility and ease of extendability, two well known design patters are adopted in Ludwig: the strategy pattern~\cite{gamma1994design} and the registry pattern.
	The strategy pattern is adopted at different levels to allow different behaviors to be performed by different instantiations of the same abstract components.
	It is used both to make the different data types interchangeable from the point of view of model building, training, and inference, and to make different encoders and decoders for the same type interchangeable.
	The registry pattern, on the other hand, is implemented in Ludwig by assigning names to code constructs (either variables, function, objects, or modules) and storing them in a dictionary.
	They can be referenced by their name, allowing for straightforward extensibility; adding an additional behavior is as simple as adding a new entry in the registry.
	
	In Ludwig, the combination of these two patterns allows users to add new behaviors by simply implementing the abstract function interface of the encoder of a specific type and adding that function implementation in the registry of implementations available.
	The same applies for adding new decoders, new combiners, and to add additional data types.
	The problem with this approach is that different implementations of the same abstract functions have to conform to the same interface, but in our case some parameters of the function may be different.
	As a concrete example, consider two text encoders: a recurrent neural network (RNN) and a convolutional neural network (CNN).
	Although they both conform to the same abstract encoding function in terms of the rank of the input and output tensors, their hyper-parameters are different, with the RNN requiring a boolean parameter indicating whether to apply bi-direction or not, and the CNN requiring the size of the filters.
	Ludwig solves this problem by exploiting \textit{**kwargs}, a Python functionality that allows to pass additional parameters to functions by specifying their names and collecting them into a dictionary.
	This allows different functions implementing the same abstract interface to have the same signature and then retrieve the specific additional parameters from the dictionary using their names.
	This also greatly simplifies the implementation of default parameters, because if the dictionary does not contain the keyword of a required parameter, the default value for that parameters is used instead automatically.

	\subsection{Training Pipeline}
	
	\begin{figure*}[h]
		\centering
		\includegraphics[width=\linewidth]{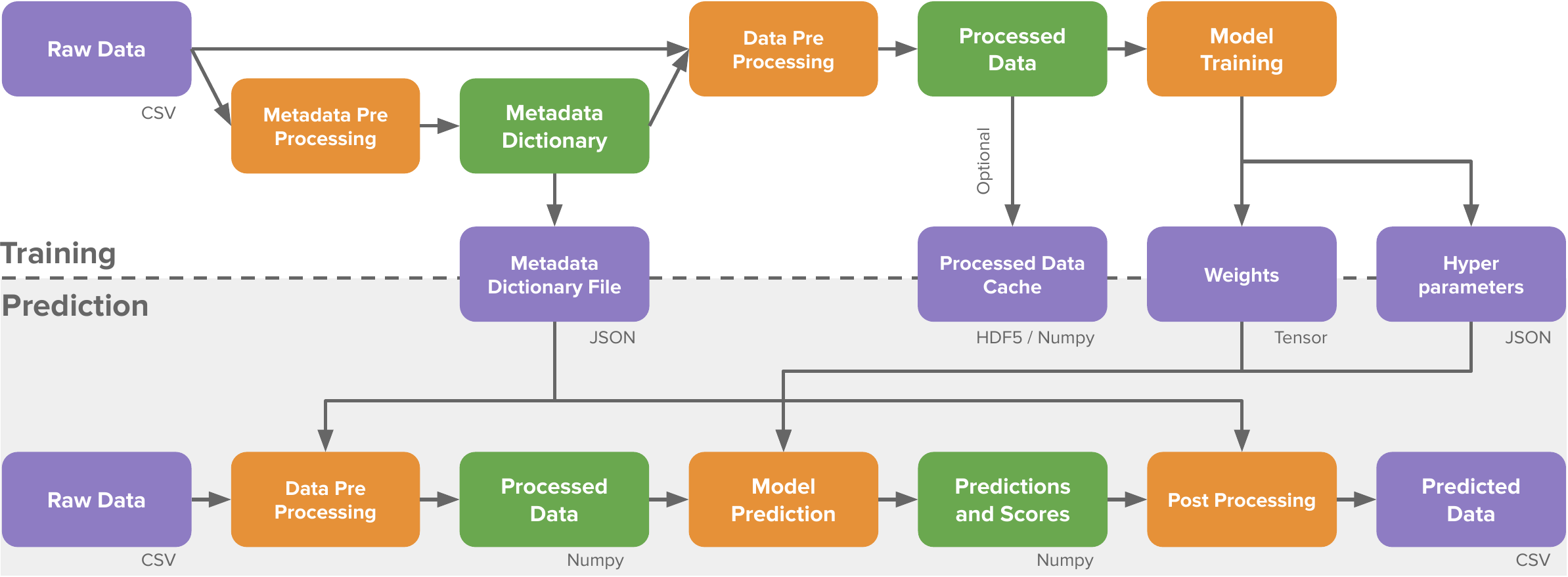}
		\caption{A depiction of the training and prediction pipeline.}
		\label{fig:pipeline}
	\end{figure*}
	
	Given a model definition, Ludwig builds a training pipeline as shown in the top of Figure~\ref{fig:pipeline}.
	The process is not particularly different from many other machine learning tools and consists in a metadata collection phase, a data pre-processing phase, and a model training phase.
	The metadata mappings in particular are needed in order to apply exactly the same pre-processing to input data at prediction time, while model weights and hyper-parameters are saved in order to load the same exact model obtained during training.
	The main notable innovation is the fact that every single component, from the pre-processing to the model, to the training loop is dynamically built depending on the declarative model definition.
	
	One of the main use cases of Ludwig is the quick exploration of different model alternatives for the same data, so, after pre-processing, the pre-processed data is optionally cached into an HDF5 file.
	The next time the same data is accessed, the HDF5 file will be used instead, saving the time needed to pre-process it.
	
	\subsection{Prediction Pipeline}
	
	The prediction pipeline is depicted in the bottom of Figure~\ref{fig:pipeline}.
	It uses the metadata obtained during the training phase to pre-process the new input data, loads the model reading its hyper-parameters and weights, and uses it to obtain predictions that are mapped back in data space by a post-processor that uses the same mappings obtained at training time.

	\section{Evaluation}

    One of the positive effects of the \textit{ECD} architecture and its implementation in Ludwig is the ability to specify a potentially complex model architecture with a concise declarative model definition.
    To analyze how much of an impact this has on the amount of code needed to implement a model (including pre-processing, the model itself, the training loop, and the metrics calculation), the number of lines of code required to implement four reference architectures using different libraries is compared: WordCNN~\cite{DBLP:conf/emnlp/Kim14},  Bi-LSTM~\cite{DBLP:conf/acl/TaiSM15} - both models for text classification and sentiment analysis, Tagger~\cite{DBLP:conf/naacl/LampleBSKD16} - sequence tagging model with an RNN encoder and a per-token classification, ResNet~\cite{he2016deep} - image classification model.
    Although this evaluation is imprecise in nature (the same model can be implemented in a more or less concise way and writing a parameter in a configuration file is substantially simpler than writing a line of code), it could provide intuition about the amount of effort needed to implement a model with different tools.
    To calculate the mean for different libraries, openly available implementations on GitHub are collected and the number of lines of code of each of them is collected (the list of repositories is available in the appendix).
    For Ludwig, the amount of lines in the configuration file needed to obtain the same models is reported both in the case where no hyper-parameter is spacified and in the case where all its hyper-parameters are specified.
    
    The results in Table~\ref{tab:evaluation} show how even when specifying all its hyper-parameters, a Ludwig declarative model configuration is an order of magnitude smaller than even the most concise alternative.
    This supports the claim that Ludwig can be useful as a tool to reduce the effort needed for training and using deep learning models.
    
    \begin{table}[]
    {\small
    \begin{tabular}{r|ccc|cc}
             & TensorFlow & Keras     & PyTorch   & \multicolumn{2}{c}{Ludwig} \\
             & mean       & mean      & mean.     & w/o  & w \\
    \hline
    WordCNN  &  406.17  &  201.50  &  458.75  &  8  & 66 \\
    Bi-LSTM  &  416.75  &  439.75  &  323.40  & 10  & 68 \\
    Tagger   & 1067.00  & 1039.25  & 1968.00  & 10  & 68 \\
    ResNet   & 1252.75  &  779.60  &  479.43  &  9  & 61
    \end{tabular}
    }
    \caption{Number of lines of code for implementing different models. \textit{mean} columns are the mean lines of code needed to write a program from scratch for the task. \textit{w} and \textit{w/o} in the Ludwig column refer to the number of lines for writing a model definition specifying every single model hyper-parameter and pre-processing parameter, and without specifying any hyper-parameter respectively.}
	\label{tab:evaluation}
    \end{table}
    
	\section{Limitations and future work}
	
	Although Ludwig's \textit{ECD} architecture is particularly well-suited for supervised and self-supervised tasks, how suitable it is for other machine learning tasks is not immediately evident.
	
	One notable example of such tasks are Generative Adversarial Networks (GANs)~\cite{goodfellow2014generative}: their architecture contains two models that learn to generate data and discriminate synthetic from real data and are trained with inverted losses.
	In order to replicate a GAN within the boundaries of the \textit{ECD} architecture, the inputs to both models would have to be defined at the encoder level, the discriminator output would have to be defined as a decoder, and the remaining parts of both models would have to be defined as one big combiner, which is inelegant; for instance, changing just the generator would result in an entirely new implementation.
	An elegant solution would allow for disentangling the two models and change them independently.
	The recursive graph extension of the combiner described in section~\ref{sub:ECD} allows a more elegant solution by providing a mechanism for defining the generator and discriminator as two independent sub-graphs, improving modularity and extensibility.
	WAn extension of the toolbox in this direction is planned in the future.
	
	Another example is reinforcement learning.
	Although \textit{ECD} can be used to build the vast majority of deep architectures currently adopted in reinforcement learning, some of the techniques they employ are relatively hard to represent, such as instance double inference with fixed weights in Deep Q-Networks~\cite{DBLP:journals/nature/MnihKSRVBGRFOPB15}, which can currently be implemented only with a really custom and inelegant combiner.
	Moreover, supporting the dynamic interaction with an environment for data collection and more clever ways to collect it like Go-Explore's~\cite{Ecoffet2019goexplore} archive or prioritized experience replay~\cite{Schaul2015prioritizedexperiencereplay}, is currently out of the scope of the toolbox: a user would have to build these capabilities on their own and call Ludwig functions only inside the inner loop of the interaction with the environment.
	Extending the toolbox to allow for easy adoption in reinforcement learning scenarios, for example by allowing training through policy gradient methods like REINFORCE~\cite{Williams92reinforce} or off-policy methods, is a potential direction of improvement.
	
	Although these two cases highlight current limitations of the Ludwig, it's worth noting how most of the current industrial applications of machine learning are based on supervised learning, and that is where the proposed architecture fits the best and the toolbox provides most of its value.
	
	Although the declarative nature of Ludwig's model definition allows for easier model development, as the number of encoders and their hyper-parameters increase, the need for automatic hyper-parameter optimization arises.
	In Ludwig, however, different encoders and decoders, i.e., sub-modules of the whole architecture, are themselves hyper-parameters.
	For this reason, Ludwig is well-suited for performing both hyper-parameter search and architecture search, and blurs the line between the two.
	
	A future addition to the model definition file will be an hyper-parameter search section that will allow users to define which strategy among those available to adopt to perform the optimization and, if the optimization process itself contains parameters, the user will be allowed to provide them in this section as well.
	Currently a Bayesian optimization over combinatorial structures~\cite{BaptistaP18BOCS} approach is in development, but more can be added.
	
	Finally, more feature types will be added in the future, in particular videos and graphs, together with a number of pre-trained encoders and decoders, which will allow training of a full model in few iterations.
	
	\section{Related Work}
	
	TensorFlow~\cite{tensorflow2015-whitepaper}, Caffe~\cite{jia2014caffe}, Theano~\cite{2016arXiv160502688short} and other similar libraries are tensor computation frameworks that allow for automatic differentiation and declarative model through the definition of a computation graph.
	They all provide similar underlying primitives and support computation graph optimizations that allow for training of large-scale deep neural networks.
	PyTorch~\cite{paszke2017automatic}, on the other hand, provides the same level of abstraction, but allows users to define models imperatively: this has the advantage to make a PyTorch program easier to debug and to inspect.
	By adding eager execution, TensorFlow 2.0 allows for both declarative and imperative programming styles.
	In contrast, Ludwig, which is built on top of TensorFlow, provides a higher level of abstraction for the user.
	Users can declare full model architectures rather than underlying tensor operations, which allows for more concise model definitions, while flexibility is ensured by allowing users to change each parameter of each component of the architecture if they wish to.
	
	Sonnet~\cite{sonnetblog}, Keras~\cite{chollet2015keras}, and AllenNLP~\cite{Gardner2017AllenNLP} are similar to Ludwig in the sense that both libraries provide a higher level of abstraction over TensorFlow and PyTorch primitives respectively.
	However, while they provide modules which can be used to build a desired network architecture, what distinguishes Ludwig from them is its declarative nature and being built around data type abstraction.
	This allows for the flexible \textit{ECD} architecture that can cover many use cases beyond the natural language processing covered by AllenNLP, and also doesn't require to write code for both model implementation and pre-processing like in Sonnet and Keras.
	
	Scikit-learn~\cite{sklearn_api}, Weka~\cite{DBLP:journals/sigkdd/HallFHPRW09}, and MLLib~\cite{meng2016mllib} are popular machine learning libraries among researchers and industry practitioners.
	They contain implementations of several different traditional machine learning algorithm and provide common interfaces for them to use, so that algorithms become in most cases interchangeable and users can easily compare them.
	Ludwig follows this API design philosophy in its programmatic interface, but focuses on deep learning models that are not available in those tools.

	\section{Conclusions}
	
	This work presented Ludwig, a deep learning toolbox built around type-based abstraction and a flexible \textit{ECD} architecture that allows model definition through a declarative language.
	
	The proposed tool has many advantages in terms of flexibility, extensibility, and ease of use, which allow both experts and novices to train deep learning models, employ them for obtaining predictions, and experiment with different architectures without the need to write code, but still allowing users to easily add custom sub-modules.
	
	In conclusion, Ludwig's general and flexible architecture and its ease of use make it a good option for democratizing deep learning by making it more accessible, streamlining and speeding up experimentation, and unlocking many new applications.


\bibliographystyle{sysml2019}
\bibliography{bibliography}

\newpage
\onecolumn
\appendix
    \section{Full list of GitHub repositories}

    \begin{table*}[h]
    {\small
    \begin{tabular}{p{.48\textwidth}ccp{.25\textwidth}}
    repository & loc & model & notes \\
    \hline
    \url{https://github.com/dennybritz/cnn-text-classification-tf} & 308 & WordCNN  & \\
    \url{https://github.com/randomrandom/deep-atrous-cnn-sentiment} & 621 & WordCNN  & \\
    \url{https://github.com/jiegzhan/multi-class-text-classification-cnn} & 284 & WordCNN  & \\
    \url{https://github.com/TobiasLee/Text-Classification} & 335 & WordCNN  & cnn.py + files in utils directory \\
    \url{https://github.com/zackhy/TextClassification} & 405 & WordCNN  & cnn\_classifier.pt + train.py + test.py \\
    \url{https://github.com/YCG09/tf-text-classification} & 484 & WordCNN  & all files minus the rnn related ones \\
    \url{https://github.com/roomylee/rnn-text-classification-tf} & 305 & Bi-LSTM  & \\
    \url{https://github.com/dongjun-Lee/rnn-text-classification-tf} & 271 & Bi-LSTM  & \\
    \url{https://github.com/TobiasLee/Text-Classification} & 397 & Bi-LSTM  & attn\_bi\_lstm.py + files utils directory \\
    \url{https://github.com/zackhy/TextClassification} & 459 & Bi-LSTM  & rnn\_classifier.pt + train.py + test.py \\
    \url{https://github.com/YCG09/tf-text-classification} & 506 & Bi-LSTM  & all files minus the cnn related ones \\
    \url{https://github.com/ry/tensorflow-resnet} & 2243 & ResNet  & \\
    \url{https://github.com/wenxinxu/resnet-in-tensorflow} & 635 & ResNet  & \\
    \url{https://github.com/taki0112/ResNet-Tensorflow} & 472 & ResNet  & \\
    \url{https://github.com/ShHsLin/resnet-tensorflow} & 1661 & ResNet  & \\
    \url{https://github.com/guillaumegenthial/sequence\_tagging} & 959 & Tagger  & \\
    \url{https://github.com/guillaumegenthial/tf\_ner} & 1877 & Tagger  & \\
    \url{https://github.com/kamalkraj/Named-Entity-Recognition-with-Bidirectional-LSTM-CNNs} & 365 & Tagger  & \\
    \end{tabular}
    }
    \caption{List of TensorFlow repositories used for the evaluation.}
	\label{tab:raw_evaluation_tensorflow}
    \end{table*}

    \begin{table*}[]
    {\small
    \begin{tabular}{p{.48\textwidth}ccp{.25\textwidth}}
    repository & loc & model & notes \\
    \hline
    \url{https://github.com/Jverma/cnn-text-classification-keras} & 228 & WordCNN  & \\
    \url{https://github.com/bhaveshoswal/CNN-text-classification-keras} & 117 & WordCNN  & \\
    \url{https://github.com/alexander-rakhlin/CNN-for-Sentence-Classification-in-Keras} & 258 & WordCNN  & \\
    \url{https://github.com/junwang4/CNN-sentence-classification-keras-2018} & 295 & WordCNN  & \\
    \url{https://github.com/cmasch/cnn-text-classification} & 122 & WordCNN  & \\
    \url{https://github.com/diegoschapira/CNN-Text-Classifier-using-Keras} & 189 & WordCNN  & \\
    \url{https://github.com/shashank-bhatt-07/Keras-LSTM-Sentiment-Classification} & 425 & Bi-LSTM  & \\
    \url{https://github.com/AlexGidiotis/Document-Classifier-LSTM} & 678 & Bi-LSTM  & \\
    \url{https://github.com/pinae/LSTM-Classification} & 547 & Bi-LSTM  & \\
    \url{https://github.com/susanli2016/NLP-with-Python/blob/master/Multi-Class\%20Text\%20Classification\%20LSTM\%20Consumer\%20complaints.ipynb} & 109 & Bi-LSTM  & \\
    \url{https://github.com/raghakot/keras-resnet} & 292 & ResNet  & \\
    \url{https://github.com/keras-team/keras-applications/blob/master/keras\_applications/resnet50.py} & 297 & ResNet  & Only model, no preprocessing \\
    \url{https://github.com/broadinstitute/keras-resnet} & 2285 & ResNet  & \\
    \url{https://github.com/yuyang-huang/keras-inception-resnet-v2} & 560 & ResNet  & \\
    \url{https://github.com/keras-team/keras-contrib/blob/master/keras\_contrib/applications/resnet.py} & 464 & ResNet  & Only model, no preprocessing \\
    \url{https://github.com/Hironsan/anago} & 2057 & Tagger  & \\
    \url{https://github.com/floydhub/named-entity-recognition-template} & 150 & Tagger  & \\
    \url{https://github.com/digitalprk/KoreaNER} & 501 & Tagger  & \\
    \url{https://github.com/vunb/anago-tagger} & 1449 & Tagger  & \\
    \end{tabular}
    }
    \caption{List of Keras repositories used for the evaluation.}
	\label{tab:raw_evaluation_keras}
    \end{table*}
    
    \begin{table*}[]
    {\small
    \begin{tabular}{p{.48\textwidth}ccp{.25\textwidth}}
    repository & loc & model & notes \\
    \hline
    \url{https://github.com/Shawn1993/cnn-text-classification-pytorch} & 311 & WordCNN  & \\
    \url{https://github.com/yongjincho/cnn-text-classification-pytorch} & 247 & WordCNN  & \\
    \url{https://github.com/srviest/char-cnn-text-classification-pytorch} & 778 & WordCNN  & ignored model\_CharCNN2d.py \\
    \url{https://github.com/threelittlemonkeys/cnn-text-classification-pytorch} & 499 & WordCNN  & \\
    \url{https://github.com/keishinkickback/Pytorch-RNN-text-classification} & 414 & Bi-LSTM  & \\
    \url{https://github.com/Jarvx/text-classification-pytorch} & 421 & Bi-LSTM  & \\
    \url{https://github.com/jiangqy/LSTM-Classification-Pytorch} & 324 & Bi-LSTM  & \\
    \url{https://github.com/a7b23/text-classification-in-pytorch-using-lstm} & 188 & Bi-LSTM  & \\
    \url{https://github.com/claravania/lstm-pytorch} & 270 & Bi-LSTM  & \\
    \url{https://github.com/hysts/pytorch\_resnet} & 447 & ResNet  & \\
    \url{https://github.com/a-martyn/resnet} & 286 & ResNet  & \\
    \url{https://github.com/hysts/pytorch\_resnet\_preact} & 535 & ResNet  & \\
    \url{https://github.com/ppwwyyxx/GroupNorm-reproduce/tree/master/ImageNet-ResNet-PyTorch} & 1095 & ResNet  & \\
    \url{https://github.com/KellerJordan/ResNet-PyTorch-CIFAR10} & 199 & ResNet  & \\
    \url{https://github.com/mbsariyildiz/resnet-pytorch} & 450 & ResNet  & \\
    \url{https://github.com/akamaster/pytorch\_resnet\_cifar10} & 344 & ResNet  & \\
    \url{https://github.com/ZhixiuYe/NER-pytorch} & 1184 & Tagger  & \\
    \url{https://github.com/sgrvinod/a-PyTorch-Tutorial-to-Sequence-Labeling} & 840 & Tagger  & \\
    \url{https://github.com/epwalsh/pytorch-crf} & 3243 & Tagger  & \\
    \url{https://github.com/LiyuanLucasLiu/LM-LSTM-CRF} & 2605 & Tagger  & \\
    \end{tabular}
    }
    \caption{List of Pytorch repositories used for the evaluation.}
	\label{tab:raw_evaluation_pytorch}
    \end{table*}
    

\end{document}